\documentclass[runningheads]{llncs}
\usepackage[T1]{fontenc}
\usepackage{graphicx,verbatim}
\usepackage{booktabs}
\usepackage{multirow}
\usepackage{adjustbox}
\usepackage{amssymb}
\usepackage{threeparttable}
\usepackage{amsmath}
\usepackage{algorithm}
\usepackage{algpseudocode}
\usepackage{caption}
\usepackage{orcidlink}
\usepackage{textcomp}
\usepackage{marvosym}
\begin{document}
\title{Small Lesions-aware Bidirectional Multimodal Multiscale Fusion Network for Lung Disease Classification}

\titlerunning{Small Lesions-aware BMMFN for Lung Disease Classification}

\author{Jianxun Yu\inst{1~\star}\orcidlink{0009-0008-4443-6443} \and Ruiquan Ge\inst{2}\thanks{Co-first authors.}\orcidlink{0000-0003-3380-7970}  \and  Zhipeng Wang\inst{2}\orcidlink{0009-0004-1856-360X}  \and \and  Cheng Yang\inst{2}\orcidlink{0009-0009-0947-656X}  \and Chenyu Lin\inst{2}\orcidlink{0009-0009-7991-1918}  \and Xianjun Fu\inst{3} \and Jikui Liu\inst{4} \textsuperscript{(\Letter)} \and
Ahmed Elazab\inst{5} \textsuperscript{(\Letter)} \and
Changmiao Wang\inst{6} 
}
\authorrunning{J. Yu et al.}
\institute{$^{1}$ Xidian University  \quad $^{2}$ Hangzhou Dianzi University \\ $^{3}$ Zhejiang College of Security Technology, School of Artificial Intelligence \\ $^{4}$ Shenzhen Polytechnic University \quad  $^{5}$ Shenzhen University \\
$^{6}$ Shenzhen Research Institute of Big Data \\
\email{liujikui007@gmail.com} \quad
\email{ahmedelazab@szu.edu.cn}
    }

\maketitle              
\begin{abstract}
The diagnosis of medical diseases faces challenges such as the misdiagnosis of small lesions. Deep learning, particularly multimodal approaches, has shown great potential in the field of medical disease diagnosis. However, the differences in dimensionality between medical imaging and electronic health record data present challenges for effective alignment and fusion. To address these issues, we propose the Multimodal Multiscale Cross-Attention Fusion Network (MMCAF-Net). This model employs a feature pyramid structure combined with an efficient 3D multi-scale convolutional attention module to extract lesion-specific features from 3D medical images. To further enhance multimodal data integration, MMCAF-Net incorporates a multi-scale cross-attention module, which resolves dimensional inconsistencies, enabling more effective feature fusion. We evaluated MMCAF-Net on the Lung-PET-CT-Dx dataset, and the results showed a significant improvement in diagnostic accuracy, surpassing current state-of-the-art methods. The code is available at \url{https://github.com/yjx1234/MMCAF-Net}.


\end{abstract}
\section{Introduction}

Computer-aided diagnosis is crucial in modern clinical applications, improving disease detection efficiency and accuracy. Deep learning methods are widely used for medical diagnosis, utilizing information from various modalities such as medical imaging and electronic health records to deliver comprehensive patient assessments \cite{jiang2024tabular}\cite{hager2023best}. However, detecting certain lesions, like squamous cell carcinoma and pulmonary embolism, remains challenging due to their small size in images, which can be easily missed by conventional models \cite{guo2024cpp}\cite{liu2024rotated}. Additionally, discrepancies between modalities complicate effective multimodal fusion for clinical decision-making. These challenges highlight the need for more robust methodologies to enhance multimodal data integration and improve diagnostic accuracy.

Multi-scale methods are effective for small lesion detection by capturing local and global information, but traditional approaches often incur high computational costs. To improve efficiency, Rahman \textit{et al.}. introduced the Multi-scale Convolutional Attention Module (MSCAM) \cite{rahman2024emcad}, which, while beneficial, is limited to 2D image processing and lacks applicability to 3D medical imaging, highlighting a gap in multi-scale feature extraction for volumetric data. Aligning features across modalities is a persistent challenge in multimodal learning. Yao \textit{et al.}. minimized Jensen–Shannon divergence for feature alignment \cite{yao2024drfuse}, while Sanjeev \textit{et al.}. used supervised contrastive loss to integrate tabular and image data \cite{sanjeev2023pecon}. However, these strategies often compromise the integrity of complex modalities to align with simpler ones, resulting in suboptimal performance and underscoring the need for more adaptive fusion techniques. Late fusion, known for its simplicity, is widely used but fails to capture complementary relationships between multimodal features \cite{huang2020multimodal}\cite{sanjeev2023pecon}. To address this, Hayat \textit{et al.}. proposed using LSTMs for aggregating multimodal representations as sequences of single-modal tokens \cite{hayat2022medfuse}, and Yao \textit{et al.}. introduced a disease-aware attention fusion module to weigh modalities based on disease profiles \cite{yao2024drfuse}. Other strategies, such as Pölsterl \textit{et al.}.'s dynamic affine feature transformation \cite{polsterl2021combining} and Reza \textit{et al.}.'s MMTM module for knowledge exchange between CNNs \cite{joze2020mmtm}, have also advanced intermediate fusion. However, they often overlook ambiguity in certain feature dimensions, which can degrade representation quality and limit downstream task effectiveness. Thus, more adaptive fusion mechanisms are necessary to enhance meaningful features while mitigating ambiguity. 

To overcome these challenges, we propose a novel framework that enables simultaneous multimodal fusion while effectively capturing small lesions in medical images. The key contributions of this work are summarized as follows: 1) We introduce an E3D-MSCA module to enhance lesion detection, particularly for small lesions. 2) We propose MSCA module designed to effectively integrate multimodal data, addressing the challenges posed by significant differences between modalities. 3) We develop a Bidirectional Scale Fusion (BSF) module, which specifically addresses the challenges of integrating features across two distinct linear scales.

\begin{figure}[h!]
\begin{center}
\vspace{-25pt}
\makebox[\textwidth][c]{
\includegraphics[width=0.9\columnwidth]{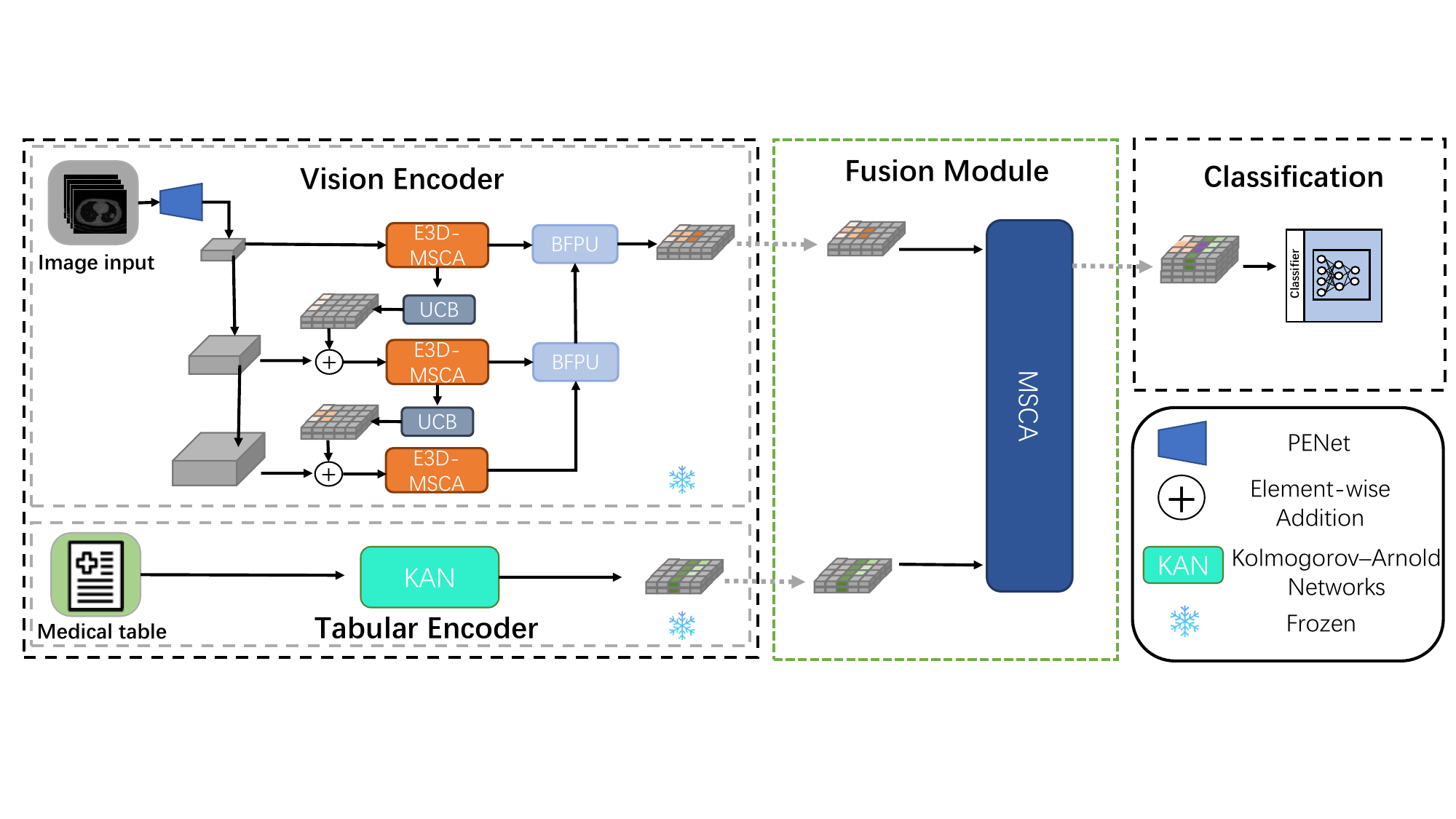}
}
\vspace{-30pt}  
\caption{Overview of the Proposed MMCAF-Net. The framework consists of an encoding module with Vision and Tabular encoders, a fusion module utilizing the MSCA module, and a classification module for final detection.
} \label{fig1}
\end{center}
\end{figure}

\section{Proposed method}
As illustrated in \textbf{Fig. \ref{fig1}}, MMCAF-Net consists of three main components. The first component is an image encoder that integrates a feature pyramid with E3D-MSCA module, designed to capture both local and global features of imaging data. This allows for the effective differentiation of challenging cases. The second component employs Kolmogorov–Arnold Networks (KAN) \cite{liu2024kan} to encode tabular features. The advantages of KAN lie in its faster scaling laws, meaning that as the model size increases, its efficiency improves significantly. Additionally, KAN demonstrates strong expressive capability even with fewer parameters. Initially, KAN was applied to tasks such as function approximation and solving partial differential equations (PDEs). Subsequent research has expanded its use to time series forecasting tasks. In addressing our problem, we flexibly apply KAN to encode our tabular data. Lastly, the third component utilizes the MSCA module to align and fuse the features from both modalities.

\begin{figure}[h!]
\begin{center}
\vspace{-5pt}
\makebox[\textwidth][c]{
\includegraphics[width=0.9\columnwidth]{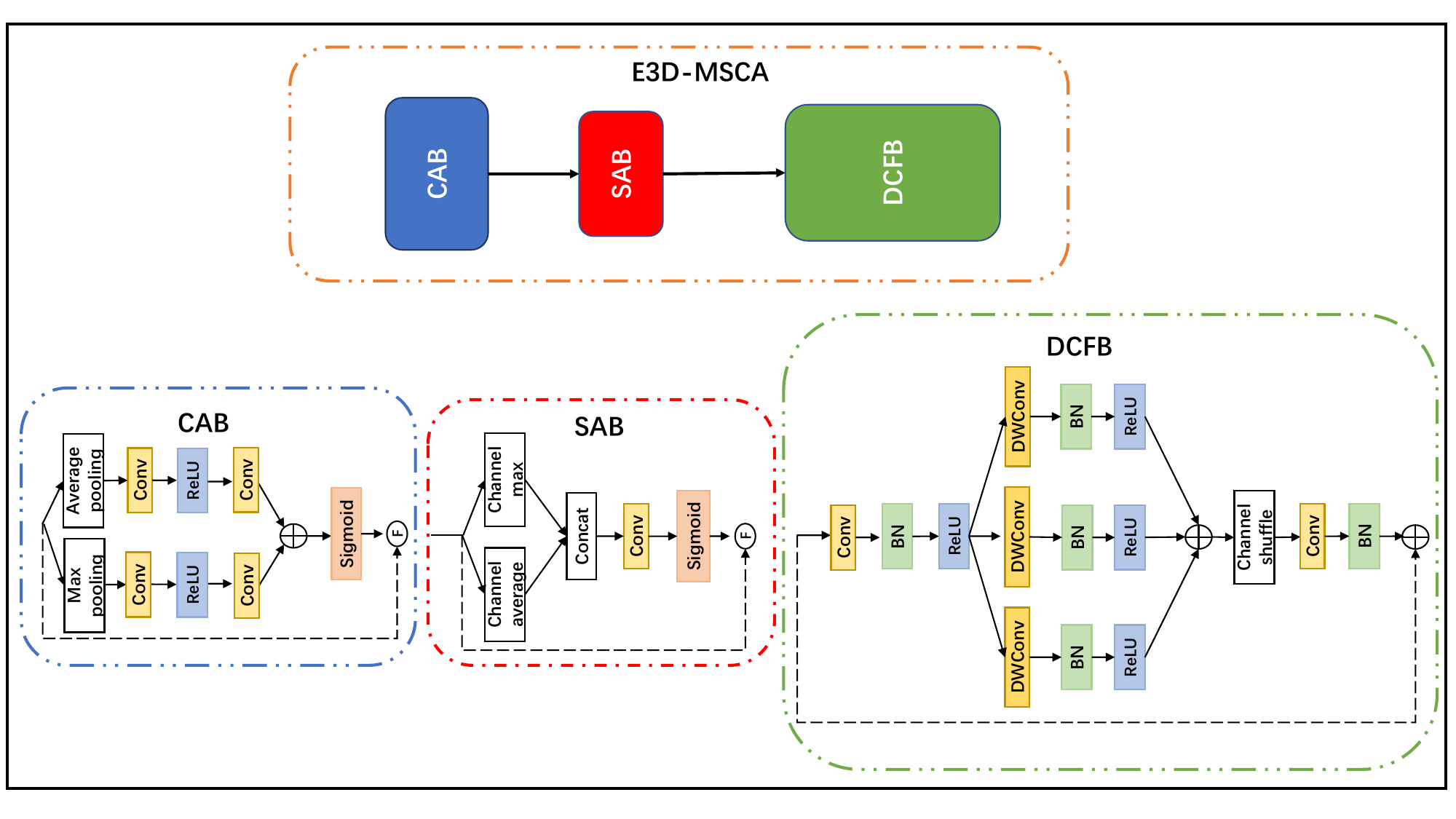}
}
\vspace{-5pt}  
\caption{Detailed structure of E3D-MSCA. It comprises 3D CAB, 3D SAB, and 3D DCFB.} \label{fig2}
\end{center}
\end{figure}

\subsection{Efficient 3D Multi-scale PENet}
Directly using the conventional lung encoder, PENet \cite{huang2020penet} has proven insufficient for effectively distinguishing challenging cases. We found that these cases are particularly difficult to differentiate due to the small size of the lesion regions, which occupy only a small portion of the entire image. Traditional encoders often overlook these lesions. Previous studies have typically employed feature pyramids to focus on both local and global features \cite{ma2023pivit}\cite{zhu2024low}. Inspired by these works, we propose an improvement based on the MSCAM method \cite{rahman2024emcad}. Specifically, we introduce a novel E3D-MSCA method, which combines the classical encoder with enhanced attention mechanisms. As shown in \textbf{Fig. \ref{fig2}}, E3D-MSCA consists of three modules: 3D Channel Attention Block(CAB), 3D Spatial Attention Block(SAB), and 3D Depth-Wise Convolution Fusion Block(DCFB). The formulation of E3D-MSCA is described as follows.
\begin{equation}
E3D-MSCA(x)=DCFB(SAB(CAB(x))),
\end{equation}
where \textit{x} represents the output features from the feature pyramid. By utilizing depth-wise convolution within the \textit{DCFB}, we significantly enhance the performance of PENet in all metrics without incurring substantial computational costs. Finally, we incorporate the Bidirectional Feedback Propagation Unit (BFPU) proposed by Chen \textit{et al.}. \cite{chen2024bidirectional} to effectively fuse features from multiple scales. The formulation of BFPU is provided using:
\begin{align}
    F_{mid} &= S\left( \mathrm{Conv}_{3\times3}(F_a) \otimes \mathrm{Conv}_{3\times3}(F_b) \right), \\
    F_{out} &= \left[ F_a + F_{mid} \otimes F_a, \, F_b + F_{mid} \otimes F_b \right],
\end{align}
where \( F_a \) and \( F_b \) represent features extracted from two different scales. The functions \( S \) and  \( \otimes \) denote the sigmoid activation function and element-wise multiplication, respectively, while the notation \( [\cdot] \) signifies channel-wise concatenation.

\subsection{Multiscale Cross Attention}
\begin{figure}[h!]
\begin{center}
\makebox[\textwidth][c]{
\includegraphics[width=0.9\columnwidth]{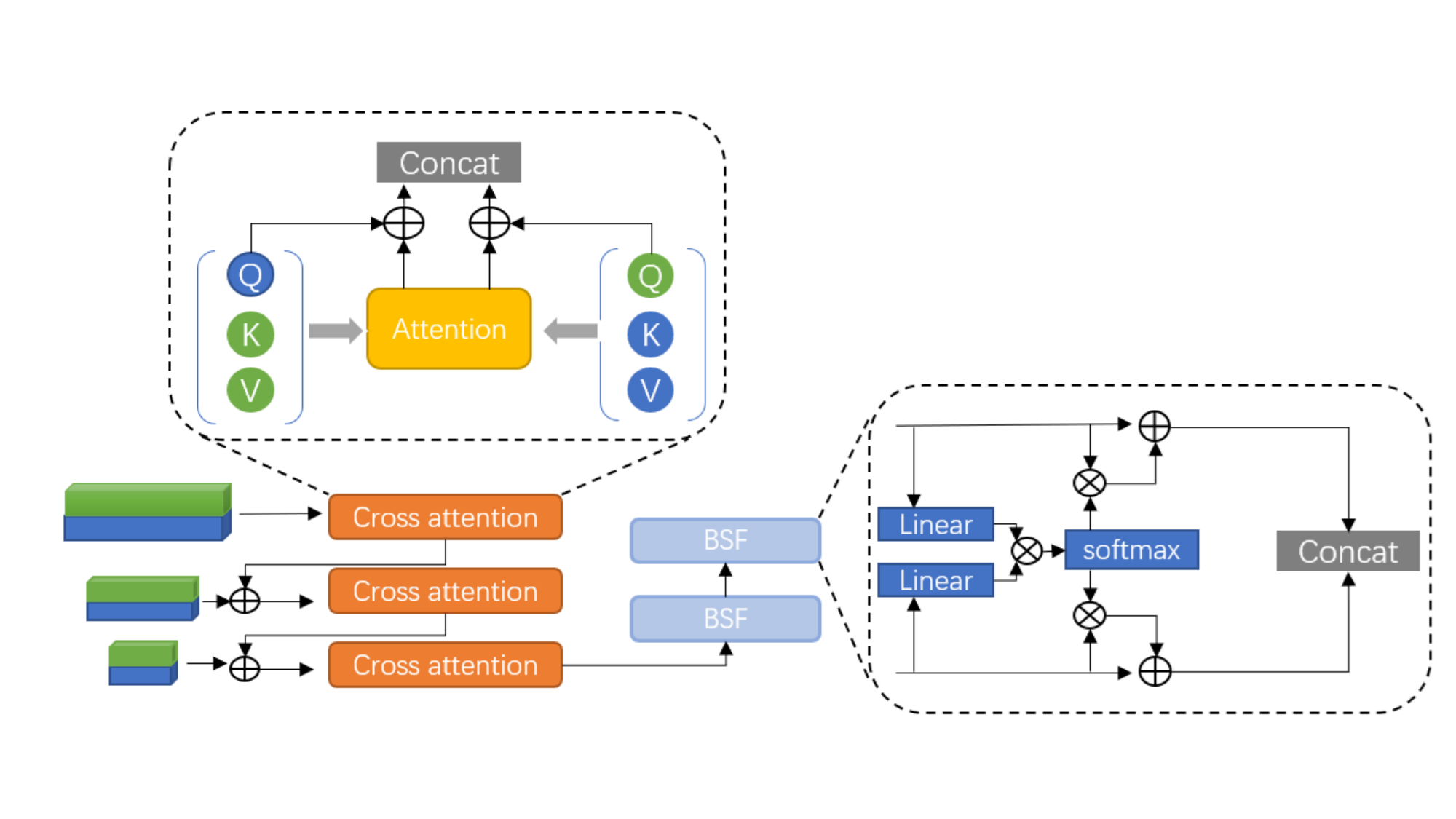}
}
\caption{Detailed structure of MSCA module.} \label{fig3}
\end{center}
\end{figure}

Due to the significant dimensional disparity between medical images and clinical data tables, we performed feature dimensionality reduction in vision encoder to align the dimensions with those of the tabular data. However, the differing distributions of image and tabular features make direct cross-fusion ineffective for leveraging the information from both sources. Moreover, dimensional conflicts can degrade the fusion performance. In prior work, Yu \textit{et al.}. utilized a multi-scale group aggregation bridge for feature fusion \cite{yu2024scnet} \cite{yu2025crisp}. Inspired by their approach, we propose a MSCA fusion module. As illustrated in \textbf{Fig. \ref{fig3}}, we applied consecutive inverted pyramid dimensionality reductions to both image and tabular features, resulting in three pairs of features with different dimensions.
Next, the feature pairs from each scale are fed into the Cross Attention module for feature fusion. In this module, we use the image and table features as queries (Q), while another feature is utilized as keys (K) and values (V). Following this, Q, K, and V undergo a multi-head operation. We compute the attention scores between the two modalities using Q and K, normalize these attention scores, and then multiply them with V to derive the final output. The details of this process are outlined as follows:
\begin{align}
Q' & = QW_q \quad (W_q \in \mathbb{R}^{D_q \times D_q}), \\
K' & = KW_k \quad (W_k \in \mathbb{R}^{D_k \times D_q}), \\
V' & = VW_v \quad (W_v \in \mathbb{R}^{D_v \times D_q}),
\end{align}

\noindent where Q has a shape of [B, N, $D_q$] while the keys K and values V have a shape of [B, M, $D_kv$].

\begin{align}
Q_h & = \text{reshape}(Q', B, H, N, C), \\
K_h & = \text{reshape}(K', B, H, M, C), \\
V_h & = \text{reshape}(V', B, H, M, C),
\end{align}

\noindent where \textit{C} represents the focus dimension of each attention head, \textit{H} denotes the number of attention heads. 
\begin{equation}
    \mathrm{Attention}(Q_h,K_h)=\frac{Q_hK_h^T}{\sqrt{C}},   \mathrm{O_h=Attention *V_h},
\end{equation}

Attention represents the attention weights, and $O_h$  denotes the final output features. Subsequently, we introduce the BSF module to integrate features across multiple scales. The structure of this module is illustrated in \textbf{Fig. \ref{fig3}}. Initially, we apply a linear transformation to align the two features to a common scale. Next, we compute the dimensional importance of both features by performing an element-wise multiplication, followed by calculating the Softmax scores. This process effectively reduces the significance of the uncertain dimensions. Finally, we merge the features from the two scales based on their dimensional importance.
\subsection{Loss Function}
We employ a binary cross-entropy loss to quantify the distance between
 the predicted outcomes and the ground truths. Mathematically, the classification loss \( \mathcal{L}_{cls} \) is expressed as follows:
\begin{equation}
\mathcal{L}_{cls}(y, \hat{y}) = -\frac{1}{N} \sum_{i=1}^{N} \left[ y_i \log(\hat{y}_i) + (1 - y_i) \log(1 - \hat{y}_i) \right],
\end{equation}
where \textit{N} represents the number of samples, $y_i$ signifies the real labels for sample i, $\hat{y}_i$ is the probability values of the model predicting sample \textit{i} range between (0,1). 

\section{Experiments}
\subsection{Dataset and Implementation}
\textbf{Dataset. }We used a publicly available dataset provided by The Cancer Imaging Archive, called Lung-PET-CT-Dx \cite{lung-pet-ct}, which includes CT or PET) scans and tabular clinical information for 355 cases. The dataset provides tumor classification labels for each patient. Most of the PET/CT data is stored in DICOM format, and all data has undergone de-identification. The dataset includes 251 adenocarcinoma samples and 61 squamous cell carcinoma samples. Since the adenocarcinoma class and squamous cell carcinoma class samples are very imbalanced in the original dataset, we decided to oversample the squamous cell carcinoma class. We use a random oversampling strategy to supplement the training set from 34 samples to 198 samples, and the number of verification sets and test sets remains the same, which are 12 and 15 respectively. The tabular data we use includes attributes such as gender, age, weight, TNM stage, and smoking history. 

\textbf{Implementation details.} We implemented MMCAF-Net on a device equipped
with Tesla V100S-PCIE-32GB using torch-2.4.1. During the training phase, we utilized binary cross-entropy loss in conjunction with the stochastic gradient descent optimizer. The model was trained for 50 epochs, with an initial learning rate set at 0.0001 and a weight decay of 0.01. We employed a batch size of 4 and processed 12 slices per input sample. Data augmentation techniques, including rotation, sharpening, and normalization, were applied to the input images, which were subsequently reshaped to a dimension of 192 × 192 pixels.

\subsection{Experiment Result}
\begin{figure}[h!]
\begin{center}
\makebox[\textwidth][c]{
\includegraphics[width=0.9\columnwidth]{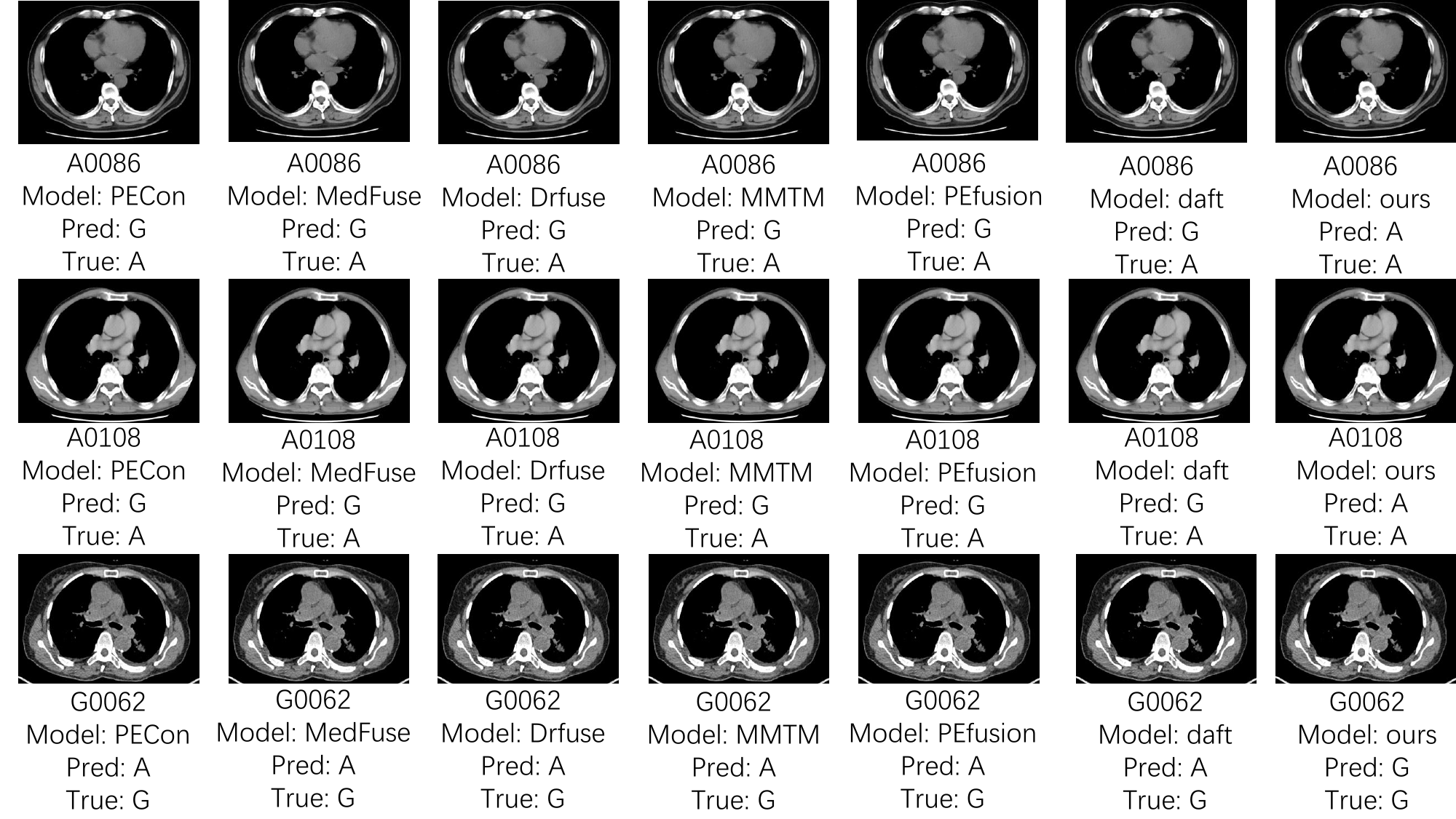}
}
\caption{Visual Representation of Classification Performance for Challenging Cases. The G denotes lung squamous cell carcinoma and A denotes lung adenocarcinoma.} \label{fig4}
\end{center}
\end{figure}

\textbf{The Lung-PET-CT Result.} In our study, we compared MMCAF-Net against the six state-of-the-art multi-modal approaches using the Lung-PET-CT dataset. As shown in \textbf{Table \ref{tab1}}, MMCAF-Net outperforms all comparative approaches across several key metrics, including accuracy (ACC), F1 score (F1), specificity, sensitivity, positive predictive value (PPV), and negative predictive value (NPV). While MMCAF-Net falls short of mmtm by 1.6\% in area under the receiver operating characteristic curve(AUROC), it surpasses mmtm by approximately 10\% in both ACC and F1 scores, and exhibits a remarkable 15\% improvement in PPV. This indicates that MMCAF-Net achieves a lower rate of false positives, demonstrating its high accuracy and reliability. Overall, it is noteworthy that all models exhibit relatively poor performance on the F1, suggesting the presence of noise or outliers in the dataset. Additionally, we visualized the classification results for challenging cases. As shown in \textbf{Fig. \ref{fig4}}, each case has 12 slices, and we selected the most representative slice of the lesion for display. The results indicate that our model outperforms the others in distinguishing subtle and difficult cases.

\textbf{Ablation study.} We conducted ablation experiments on the Lung-PET-CT dataset to evaluate the effectiveness of image encoding and multi-modal fusion techniques. As demonstrated in \textbf{Table \ref{tab2}}, we compared our proposed Efficient 3D Multi-scale PENet with the Segment Anything Model (SAM) \cite{kirillov2023segment} combined with multi-scale techniques and the original PENet. The results indicate that our method outperforms the other two approaches across all evaluated metrics, achieving improvements of 10\% and 15\% in AUROC, and 10\% and 14\% in F1 scores, respectively. Furthermore, our method shows significant enhancements in both AUROC and NPV compared to the standalone PENet, suggesting that the improved architecture enhances the reliability of negative class sample identification. Additionally, we performed ablation experiments on different fusion methods, as presented in \textbf{Table \ref{tab3}}, comparing our proposed MSCA with Cross-Attention, Late\_Fusion, and clip aligned fusion 
 \cite{radford2021learning} techniques. The results illustrate that MSCA outperforms the other three fusion methods across most metrics, with an increase in ACC of 5\%, 10\%, and 12\%, respectively. This indicates that our fusion method demonstrates superior overall performance compared to the other approaches.

\begin{table}[t]
\centering
\caption{Quantitative comparison with other methods on the public dataset. The best and the second-best results are bolded and underlined, respectively.}
\label{tab1}
\resizebox{\textwidth}{!}{
    \begin{tabular}{@{\hspace{2pt}}c@{\hspace{4pt}}c@{\hspace{4pt}}c@{\hspace{4pt}}c@{\hspace{4pt}}c@{\hspace{4pt}}c@{\hspace{4pt}}c@{\hspace{4pt}}c@{\hspace{2pt}}}
    \toprule
        Method & AUROC $\uparrow$ & ACC $\uparrow$ & F1 $\uparrow$ & Specificity $\uparrow$ & Sensitivity $\uparrow$ & PPV $\uparrow$ & NPV $\uparrow$\\
    \midrule
        PECon~\cite{sanjeev2023pecon} & 0.786 & 0.744 & 0.645 & 0.786 & \underline{0.667} & 0.625 & \underline{0.815}\\ 
        MedFuse~\cite{hayat2022medfuse} &  0.786 & 0.721 & 0.571 & \underline{0.821} & 0.533 & 0.615 & 0.767\\
        Drfuse~\cite{yao2024drfuse} & 0.613 & 0.676 & 0.353 & 0.800 & 0.333 & 0.375 & 0.769\\
        MMTM~\cite{joze2020mmtm} & \textbf{0.802} & 0.698 & 0.581 & 0.750 & 0.600 & 0.562 & 0.778\\
        PEfusion~\cite{huang2020multimodal} & 0.740 & 0.721 & 0.600 & 0.786 & 0.600 & 0.600 & 0.786\\
        daft~\cite{polsterl2021combining} & 0.727 & \underline{0.729} & \underline{0.667} & 0.786 & 0.650 & \underline{0.684} & 0.759\\
        \textbf{MMCAF-Net (Ours)} & \underline{0.786} & \textbf{0.791} & \textbf{0.690} & \textbf{0.857} & \textbf{0.667} & \textbf{0.714} & \textbf{0.828}\\
    \bottomrule
    \end{tabular}
}
\end{table}

\begin{table}[t]
\centering
\caption{Image encoder ablation experiments on the Lung-PET-CT dataset. The best and second-best results are bolded and underlined, respectively.}
\label{tab2}
\resizebox{\textwidth}{!}{
    \begin{tabular}{@{\hspace{2pt}}c@{\hspace{4pt}}c@{\hspace{4pt}}c@{\hspace{4pt}}c@{\hspace{4pt}}c@{\hspace{4pt}}c@{\hspace{4pt}}c@{\hspace{4pt}}c@{\hspace{2pt}}}
    \toprule
        Method & AUROC $\uparrow$ & ACC $\uparrow$ & F1 $\uparrow$ & Specificity $\uparrow$ & Sensitivity $\uparrow$ & PPV $\uparrow$ & NPV $\uparrow$\\
    \midrule
        SAM+E3D-MSCA & 0.610 & 0.651 & 0.444 & 0.786 & 0.400 & 0.500 & 0.710\\ 
        PENet~\cite{huang2020penet} &  0.560 & 0.721 & 0.400 & \underline{0.964} & 0.267 & \underline{0.800} & 0.711\\
        PENet+E3D-MSCA & \underline{0.644} & \underline{0.721} & \underline{0.500} & 0.893 & \underline{0.400} & 0.667 & \underline{0.735}\\
        PENet+E3D-MSCA+drop & \textbf{0.712} & \textbf{0.767} & \textbf{0.545} & \textbf{0.964} & \textbf{0.400} & \textbf{0.857} & \textbf{0.857}\\
    \bottomrule
    \end{tabular}
}
\end{table}

\begin{table}[t]
\centering
\caption{Fusion ablation experiments on the Lung-PET-CT dataset. The best and second-best results are bolded and underlined, respectively.}
\label{tab3}
\resizebox{\textwidth}{!}{
    \begin{tabular}{@{\hspace{2pt}}c@{\hspace{4pt}}c@{\hspace{4pt}}c@{\hspace{4pt}}c@{\hspace{4pt}}c@{\hspace{4pt}}c@{\hspace{4pt}}c@{\hspace{4pt}}c@{\hspace{2pt}}}
    \toprule
        Method & AUROC $\uparrow$ & ACC $\uparrow$ & F1 $\uparrow$ & Specificity $\uparrow$ & Sensitivity $\uparrow$ & PPV $\uparrow$ & NPV $\uparrow$\\
    \midrule
        Cross-Attention & \underline{0.752} & \underline{0.744} & \underline{0.686} & 0.714 & \textbf{0.800} & 0.600 & \textbf{0.870}\\ 
        clip\_Fusion~\cite{radford2021learning} &  0.717 & 0.698 & 0.629 & 0.679 & \underline{0.733} & 0.550 & 0.826\\
        Late\_Fusion &  0.729 & 0.674 & 0.462 & \underline{0.821} & 0.400 & 0.545 & 0.719\\
        MSCA\_Fusion & \textbf{0.786} & \textbf{0.791} & \textbf{0.690} & \textbf{0.857} & 0.667 & \textbf{0.714} & \underline{0.828}\\
    \bottomrule
    \end{tabular}
}
\end{table}

\section{Conclusion}
In this study, we introduce MMCAF-Net, a multimodal multiscale fusion model that effectively focuses on small lesions. MMCAF-Net incorporates the E3D-MSCA module to dynamically adjust feature weights across different scales, enhancing attention to small lesions. The MSCA module addresses the dimensional conflicts that arise when merging two modalities with significantly different characteristics. The BSF module tackles the impact of dimensional uncertainty between different scales on the fusion results. Comparative studies demonstrate that MMCAF-Net outperforms competing methods on most metrics, indicating its practical effectiveness. Future plans include adapting the model for the diagnosis of lung and other medical diseases in clinical settings.

\begin{credits}
\subsubsection{\ackname} This work was supported by the Open Project Program of the State Key Laboratory of CAD \& CG (No. A2410), Zhejiang University; Zhejiang Provincial Natural Science Foundation of China (No. LY21F020017, 2023C03090); National Natural Science Foundation of China (No. 61702146, 62076084, U20A20386, U22A2033); Shenzhen major science and technology project (KCXFZ20240903093923031), Special projects fund in key areas of the Guangdong Provincial Department of Education (2023ZDZX2085);  Guangdong Basic and Applied Basic Research Foundation (No. 2022A1515110570), Futian Healthcare Research Project (No.FTWS002); Medical Scientific Research Foundation of the Guangdong Province of China (A2023158,C2023106).

\subsubsection{\discintname} The authors declare no competing interests.
\end{credits}

\bibliographystyle{splncs04} 
\bibliography{paper-2709}
\end{document}